\documentclass[11pt,a4paper]{article}
\usepackage[hyperref]{naaclhlt2019}
\usepackage{times}
\usepackage{latexsym}

\usepackage{url}

\usepackage{xcolor}

\usepackage{color}
\usepackage{multirow}
\usepackage[normalem]{ulem}
\usepackage{subcaption}
\usepackage{amsmath,amsfonts,amssymb}
\usepackage{mathtools}
\usepackage{booktabs}
\usepackage{acronym}
\usepackage{microtype}

\usepackage{textcomp}

\aclfinalcopy 
\def\aclpaperid{389} 

\acrodef{VAD}[VAD]{Valence-Arousal-Dominance}
\acrodef{VA}[VA]{Valence-Arousal}
\acrodef{MMI}[MMI]{Maximum Mutual Information}
\acrodef{GRU}[GRU]{Gated Recurrent Unit}
\acrodef{LSTM}[LSTM]{Long Short Term Memory}
\acrodef{EMOTICONS}[EMOTICONS]{EMOTIonal CONversational System}
\acrodef{WI}[WI]{Word-Level Implicit Model}
\acrodef{WE}[WE]{Word-Level Explicit Model}
\acrodef{SEE}[SEE]{Sequence-Level Explicit Encoder Model}
\acrodef{SED}[SED]{Sequence-Level Explicit Decoder Model}
\acrodef{cornell}[Cornell]{Cornell Movie-Dialogs Corpus}

\usepackage{xspace}
\newcommand{\eg}{e.g.,\xspace}

\newcommand{\ie}{i.e.,\xspace}

\newcommand{\sect}{\textsection}
\newcommand{\fig}{Figure\xspace}

\DeclareMathOperator*{\argmax}{argmax}

\DeclareMathOperator*{\softmax}{softmax}

\newcommand{\norm}[1]{\left\lVert#1\right\rVert}

\title{Affect-Driven Dialog Generation}

\author{ Pierre Colombo$^{1*}$, 
  Wojciech Witon$^{1}$\thanks{\quad Both authors contributed equally to this work.} \ , 
  Ashutosh Modi$^{1}$, 
  James Kennedy$^{1}$, 
  Mubbasir Kapadia$^{1,2}$ \\
  {$^1$Disney Research, } 
  {$^{2}$Rutgers University}\\
  {\tt \{pierre.colombo, wojtek.witon\}@disneyresearch.com}  \\
  {\tt ashutosh.modi@disneyresearch.com}  \\
  {\tt james.kennedy@disneyresearch.com}  \\
  {\tt mubbasir.kapadia@rutgers.edu}
\\}
  
\date{}

\begin{document}
\maketitle

\begin{abstract}
The majority of current systems for end-to-end dialog generation focus on response quality without an explicit control over the affective content of the responses. In this paper, we present an affect-driven dialog system, which generates emotional responses in a controlled manner using a continuous representation of emotions. The system achieves this by modeling emotions at a word and sequence level using: (1) a vector representation of the desired emotion, (2)~an affect regularizer, which penalizes neutral words, and (3) an affect sampling method, which forces the neural network to generate diverse words that are emotionally relevant. During inference, we use a re-ranking procedure that aims to extract the most emotionally relevant responses using a~human-in-the-loop optimization process. We study the performance of our system in terms of both quantitative (BLEU score and response diversity), and qualitative (emotional appropriateness) measures.
\end{abstract}


\section{Introduction} \label{sec:Introduction}
Recent breakthroughs in deep learning techniques have had an impact on end-to-end conversational systems \cite{chen2017survey}. Current research is mainly focused on functional aspects of conversational systems: keyword extraction, natural language understanding, and pertinence of generated responses \cite{goal-oriented-chatbot}. Although these aspects are indeed key features for building a commercial system, most existing solutions lack social intelligence. Conversational systems could benefit from incorporating social intelligence by: (1) avoiding interaction problems that may arise when the system does not understand the user's request (\eg inappropriate responses that cause user anger) \cite{maslowskiwild}, and (2) building rapport with the user \cite{human_robot_relation}. Our method makes such conversational systems more social by outputting responses expressing emotion in a controlled manner, without sacrificing grammatical correctness, coherence, or relevance.

\begin{table*}
\small\addtolength{\tabcolsep}{0pt}
    \centering
    \begin{tabular}{rclll}
        \cmidrule{2-5}
        &Input & Good to see you again.&Be careful, I have a knife.&You're the only one who can help us.\\ \cmidrule{2-5}
        &Baseline & It's good to see you.&Don't worry about it.&There's a lot of people here.\\
        \multirow{4}{*}{\rotatebox{90}{\scriptsize EMOTICONS}}&Fear &What are you doing here?&I'm just trying to get out of here.&What are you going to do with me?\\
        &Anger &I'm not here.&I'm going to kill you.&You're not the only one who can help me.\\
        &Joy & Good to see you.&I'm going to marry you.&I can protect you.\\
        &Surprise & You want to talk to me?&I'm just going to the party.&You've got to be kidding me.\\\cmidrule{2-5}
    \end{tabular}
    \caption{\textbf{Example responses} from the baseline (seq2seq) model and the four EMOTICONS models with different emotions.}
    \label{tab:examples}
\end{table*}
Existing sequence-to-sequence (seq2seq) architectures, either recurrent- \cite{seq2seq-sordoni,seq2seq-serban}, attention- \cite{attention-is-all-you-need} or convolutional neural network (CNN)-based \cite{conv-hierarchical}, do not provide a straightforward way to generate emotionally relevant output in a controlled manner. We introduce \ac{EMOTICONS}, which generates emotion-specific responses. It is based on novel contributions presented in this paper which fall in two main categories: explicit models which allow a controlled emotion-based response generation (\eg methods based on emotion embeddings, affective sampling, and affective re-ranking), and implicit models with no direct control over the desired emotion (\ie affective regularizer). We show that \ac{EMOTICONS} outperforms both the system proposed by \citet{aaai_emotion} (current state of the art for our task) and the vanilla seq2seq in terms of BLEU score \cite{bleu} (improvement up to $7.7\%$) and response diversity (improvement up to $52\%$). Additionally, we qualitatively evaluate the emotional content of the generated text (see example responses in Table~\ref{tab:examples}). The user study (22 people) demonstrates that \ac{EMOTICONS} is able to generate grammatically correct, coherent, emotionally-rich text in a controlled manner.




\section{Related Work} \label{sec:Related_Work}
Sequence-to-sequence (seq2seq) models have attracted a lot of attention in the past few years, especially in the fields of Neural Machine Translation \cite{seq2seq,attention-original} and Neural Dialogue Generation \cite{seq2seq-sordoni,neural-conv-model,seq2seq-serban}. Prior work has focused on designing architectures that lead to the best performance in terms of BLEU \cite{bleu} and Perplexity scores. Most seq2seq models are based on gated recurrent neural networks, either \ac{LSTM} \cite{lstm} or \ac{GRU} \cite{seq2seq-serban}, but in general it is difficult to conclude which gating mechanism performs better \cite{gru-lstm-comparison}. In our model, we use \ac{GRU} because it has fewer parameters to optimize, and it is faster to train.

In order to overcome the problem of generating trivial or mundane responses, there have been developments in inference techniques for encoder-decoder systems. Use of beam search has been shown to improve the general quality of generated answers, while \ac{MMI} \cite{mmi} has improved the diversity of generated answers, leading to more meaningful output. We build on these techniques during affective inference.

Emotion-based (affective) dialog generation systems have received increasing attention in the past few years. \newcite{emotion-token} use emotion tokens (special ``words'' in a dictionary representing specific emotions) at either the encoder or decoder side, forcing the decoder to output a sentence with one specific emotion. \newcite{aaai_emotion} build their system using external and internal memory, where the former forces the network to generate emotional words, and the latter measures how emotional a generated sequence is compared to a target sequence. \newcite{affective-generation} modeled emotions in \ac{VA} space for response generation. We extend this idea by using a \ac{VAD} Lexicon \cite{vad_dictionary}, as it has been shown by \newcite{vad_usage} that the third dimension (Dominance) is useful for modeling affect. \newcite{ashgar} used the \ac{VAD} Lexicon, but they let the neural network choose the emotion to generate (by maximizing or minimizing the affective dissonance) and their system cannot generate different emotional outputs for the same input, nor generate a specified emotion.


\section{System Architecture} \label{sec:General_Model}
\begin{figure*}
\centering
  \includegraphics[width=\linewidth]{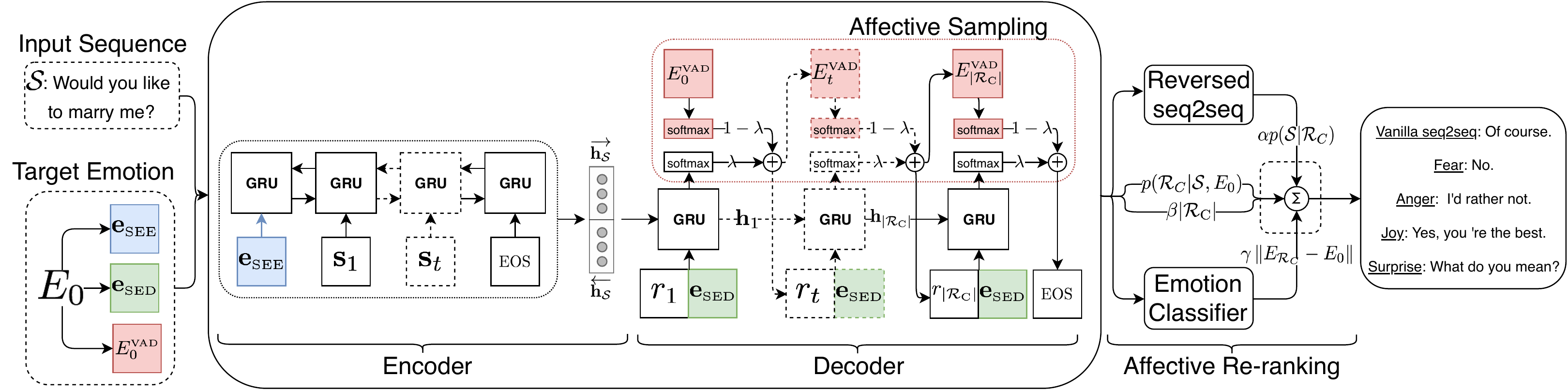}
  \caption{\textbf{System overview.} The input sequence and the target emotion (automatically extracted from the target response using the emotion classifier during training, and set by the user during inference) are fed into the seq2seq. The generated candidate responses are re-ranked based on the output of the reversed seq2seq, the length, and the emotional content.}
  \label{fig:overview}
  \vspace{-1.0em}
\end{figure*}

Our system (see overview in \fig~\ref{fig:overview}) is divided into three main components: (1) Emotion Labeling -- automatic labeling of sentences according to the emotional content they express, using an emotion classifier (\sect\ref{sec:emotion-classifier}); labeling of words with \ac{VAD} Lexicon values (\sect\ref{sec:affective-dictionary}), (2) Affective Training -- training of two seq2seq networks, which use an encoder-decoder setting. The first network is trained with prompt-response pairs (S-T), whereas the second (used during Affective Inference) is trained with reversed pairs (T-S), (3) Affective Inference -- generation of many plausible responses, which are re-ranked based on emotional content.
\subsection{Preliminaries}

Let $V=\{w_1, w_2, \dots, w_{|V|}\}$ be a vocabulary, and $\mathcal{X} = (x_1,x_2, \dots,x_{|\mathcal{X}|})$ a~sequence of words (e.g. a sentence). We denote ${E_{\mathcal{X}}} \in \mathbb{R}^{6}$ as an emotion vector representing a probability distribution over six emotions associated with the sequence~$\mathcal{X}$:
\begin{equation*}
    \label{eq:WE3}
    E_\mathcal{X} = \left[ \begin{array}{c}
        p_{\small \text{anger}} \\
        p_{\small \text{surprise}}\\ 
        p_{\small \text{joy}}\\
        p_{\small \text{sadness}}\\
    p_{\small \text{fear}}\\ 
    p_{\small \text{disgust}}
    \end{array}\right]
\end{equation*}

Note that in this work we focus on six basic emotions proposed by Paul Ekman \cite{sixemotions} but the techniques we develop are general and can  be extended to a more fine grained list of emotions.  
$\mathcal{X}$ can be an input sequence, candidate response, final response, or target response (denoted respectively as $\mathcal{S}$, $\mathcal{R}_C$, $\mathcal{R}_\text{final}$, $\mathcal{R}_0$). We introduce $E_0$, which during training, is the representation of the emotion of the target response ($\mathcal{R}_0$). During testing, $E_0$ indicates a desired emotion for the final response ($\mathcal{R}_\text{final}$), and can be set manually. For example, in the case of `anger', $E_0$ would be a one-hot vector with $1$ at the first position, and $0$ elsewhere.

In our work, we extend the standard seq2seq model \cite{seq2seq}, that predicts the final response 
$\mathcal{R}_\text{final} = \argmax_{\mathcal{R}_C} p(\mathcal{R}_C|S)$. The proposed affective system aims to extend the inference mechanism by incorporating emotions encoded in $E_0$:
\begin{equation} \label{eq:affective_conversational_task}
	\mathcal{R}_\text{final} = \argmax_{\mathcal{R}_{C}} p(\mathcal{R}_{C}|S,{E}_{0})
\end{equation} 



\subsection{Affect Modeling}
\label{sec:Emotion_Modelling}
We extend the standard seq2seq architecture by including emotion-specific information during the training and the inference. A critical challenge in both generating and evaluating responses is a reliable assessment of emotional state. We use two representations of emotion: (1) a categorical representation with six emotions (anger, surprise, joy, sadness, fear, disgust), 
and (2) a continuous representation in a \ac{VAD} space. The latter uses a \ac{VAD} Lexicon introduced by \newcite{vad_dictionary}, where each of 20k words is mapped to a 3D vector of \ac{VAD} values, ranging from 0 (lowest) to 1 (highest) ($\mathbf{v} \in [0,1]^3$). Valence measures the positivity/negativity, Arousal the excitement/calmness, and Dominance the powerfulness/weakness of the emotion expressed by a word. This expands the work of \newcite{affective-generation}, who modeled emotions only in \ac{VA} space. In the following sections we describe different versions of the proposed model.

\subsubsection{Emotion Classifier}
\label{sec:emotion-classifier} Affective training requires $E_0$, the emotion representation of the target sequence. In order to label all sentences of the corpus with $E_0$, we use an Emotion Classifier by \newcite{emotion-classifier}. The classifier predicts a probability distribution over class of six emotions. The classifier predictions for \ac{cornell} have been shown to be highly correlated with human predictions \cite{emotion-classifier}.


\subsubsection{\ac{SEE}} \label{sec:SEE} To explicitly generate responses with emotion, this version of the model includes an emotion embedding at the encoder side. We feed the encoder with $\mathcal{S'} =(\mathbf{e}_{\text{SEE}},s_{1},s_{2},\dots,s_{|S|})$, where $\mathbf{e}_{\text{SEE}} = \mathbf{A}_{SEE} E_0$ is an Emotion Embedding ($\mathbf{e}_{\text{SEE}} \in \mathbb{R}^{3}$), and $\mathbf{A}_{SEE} \in \mathbb{R}^{3\times6}$ is a mapping (learned during training) from $E_0$ into an emotion embedding space.

\subsubsection{\ac{SED}} \label{sec:SED} Another way of forcing an emotional output is to explicitly indicate the target emotion at every step in decoding along with other inputs. 
Formally, the GRU hidden state at time~$t$ is calculated as $\mathbf{h}_t=f(\mathbf{h}_{t-1}, r_t^\prime)$ with $r_{t}^\prime = \left[ r_{t-1};\mathbf{e}_{\text{SED}} \right]$, where $\mathbf{e}_{\text{SED}}$ is defined similarly as $\mathbf{e}_{\text{SEE}}$. It is worth noting that $\mathbf{A}_{SEE}$ and $\mathbf{A}_{SED}$ are different, which implies that the emotion embedding spaces they map to are also different. Compared to a similar approach introduced by \newcite{emotion-token}, our solution enables the desired emotional content, $E_0$, to be provided in a continuous space.

\subsubsection{\ac{WI}} \label{sssec:WI}  To model the word-level emotion carried by each sequence, we introduce an Affective Regularizer (AR), which expresses the affective distance between $\mathcal{R}_\text{final}$ and $\mathcal{R}_0$, in the \ac{VAD} space. It forces the neural network to prefer words in the vocabulary that carry emotions in terms of \ac{VAD}.  Mathematically, we extend the regular Negative Log Likelihood (NLL) loss with an affective regularizer, $\mathcal{L}_\text{AR}$:
\begin{equation*} \label{eq:vad_loss}
\begin{split}
\mathcal{L}  &= \mathcal{L}_\text{NLL} + \mathcal{L}_\text{AR}   \\
& = -\log p(\mathcal{R}_\text{final}|\mathcal{S}) + \mu \mathcal{L}_{\text{VAD}}(\mathcal{R}_\text{final},\mathcal{R}_0)
\end{split}
\end{equation*}
\begin{equation*}
\label{eq:affective-regularizer}
    \mathcal{L}_{\text{VAD}}(\mathcal{R}_\text{final},\mathcal{R}_0) = \norm{ \sum\limits_{t=1}^{|\mathcal{R}_\text{final}|}\frac{\mathbf{E}^\text{VAD}\mathbf{s}_t}{|\mathcal{R}_\text{final}|}  - 
\sum\limits_{t=1}^{|\mathcal{R}_0|}\frac{\mathbf{e}^\text{VAD}_{r_{0_t}}}{|\mathcal{R}_0|}}\text{,}
\end{equation*}
where $\mathbf{s}_t = \softmax(\mathbf{h}_t)$ ($\mathbf{s}_t \in \mathbb{R}^{|V|}$) is a confidence of the system of generating words $w_{1},\dots,w_{|V|}$ at time~$t$ and $\mu \in \mathbb{R}$. $\mathbf{e}^\text{VAD}_x \in \mathbb{R}^{3}$ is a~3D vector representing emotion associated with a word $x$ in VAD space (note that $\mathbf{e}^\text{VAD}_x$ is constant with respect to $t$), and $\mathbf{E}^\text{VAD} \in \mathbb{R}^{3\times |V|}$ is a matrix containing $\mathbf{e}^\text{VAD}_{w_v}$ for all $|V|$ words in the vocabulary:
\begin{equation*}
    \mathbf{E}^\text{VAD} =
    \left[\ \mathbf{e}_{w_1}^{\text{VAD}}\ ;\ \dots\ ;\ \mathbf{e}_{w_{|V|}}^{\text{VAD}}\ \right]
\end{equation*}
Intuitively, the regularizer penalizes the deviation of the emotional content of the generated response, $\mathcal{R}_\text{final}$, from the desired response, $\mathcal{R}_0$. The emotional information carried by $\mathcal{R}_\text{final}$ is the weighted sum of emotion representations $\mathbf{e}^\text{VAD}_{w_{i}}$ for all words $w_{i}$ in the vocabulary, where the weights are determined by the confidence $\mathbf{s}_t$.

\subsubsection{\ac{WE}}
\label{sec:WE} Sequential word generation allows sampling of the next word, based on the emotional content of the current incomplete sequence. If some words in a sequence do not express the target emotion $E_0$, other words can compensate for this by changing the final affective content, \eg in a sentence ``I think that the cat really loves me!'', the first 6 words are neutral, whereas the end of the sentence make it clearly express joy. We incorporate this observation by explicitly generating the next word using an Adaptive Affective Sampling Method:

\begin{align*}
    \label{eq:WE1}
    \log p(\mathcal{R}_C|\mathcal{S},E_0) &=\\
     &\sum\limits_{t=1}^{|{\mathcal{R}_\text{C}}|} \log p (r_{t} |r_{<t}, \mathbf{e}_{r_{<t}}, \mathbf{h}_{\mathcal{S}},E_0),
\end{align*}

\begin{equation*}
\begin{split}
    \label{eq:WE2}
    p (r_{t} |r_{<t}, \mathbf{e}_{r_{<t}}, \mathbf{h}_{\mathcal{S}}, E_0) &= \lambda\softmax g(\mathbf{h}_t) \ +
    \\& (1-\lambda)\softmax v(E^\text{VAD}_t)\text{,}
\end{split}
\end{equation*}



where $g(\mathbf{h}_t)$ is a linear mapping from \ac{GRU} hidden state $\mathbf{h}_t$ to an output vector of size $|V|$, and $0\leq\lambda\leq1$ is learned during training. The first term in Equation~\ref{eq:WE2} is responsible for generating words according to a language model preserving grammatical correctness of the sequence, whereas the second term forces generation of words carrying emotionally relevant content. $E_t^\text{VAD} \in \mathbb{R}^{3}$ is a vector representing the remaining emotional content needed to match a goal ($E_0^\text{VAD}$) after generating all words up to time $t$. It is updated every time a new word $r_t$ with an associated emotion vector $\mathbf{e}_{r_t}^\text{VAD}$ is generated:
\begin{equation*}
    \label{eq:WE4}
    \begin{split}
    E_t^\text{VAD} &= E_{t-1}^\text{VAD} - \mathbf{e}_{r_{t-1}}^\text{VAD}\\
    E_0^\text{VAD} &= \left\{ \begin{array}{ll}
         \sum\limits_{t=1}^{|\mathcal{R}_0|}\mathbf{e}_{{r_0}_t}^\text{VAD} \text{,}&\text{training}\\\\
         \mathbf{M}_\text{VAD}E_0 \cdot \text{max}_\text{length}\text{,}&\text{inference}
    \end{array}\right.
    \end{split}
\end{equation*}
where $\mathbf{e}_{{r_0}_t}^\text{VAD}$ is an emotion vector associated with words $r_{0_t}$ in the target sequence, $\text{max}_\text{length}$ is a maximum length set for the seq2seq model, and $\mathbf{M}_\text{VAD} \in \mathbb{R}^{3\times6}$ is a mapping from six-dimensional emotion space into \ac{VAD} space (every emotion has a \ac{VAD} vector as introduced by \newcite{mapping}, scaled to a~range [0, 1]):
\begin{equation*}
    \begin{array}{cccccccccc}
        &&\rotatebox{90}{anger}&\rotatebox{90}{surprise}&\rotatebox{90}{joy}&\rotatebox{90}{sadness}&\rotatebox{90}{fear}&\rotatebox{90}{disgust}\\
        &\multirow{3}{*}{\Bigg[}&0&1&1&0&0&0&\multirow{3}{*}{\Bigg]}&V\\
        \mathbf{M}_\text{VAD} = &&1&1&1&0&1&0.5&&A\\
        &&1&0&1&0&0&0.5&&D\\
    \end{array}
\end{equation*}
$v(E_t^\text{VAD})$ is a vector, whose i-th component measures the potential remaining emotional content of the sequence in the case of choosing the i-th word $w_i$:
\begin{equation*}
    \label{eq:WE3}
    v(E_t^\text{VAD}) = -\left[ \begin{array}{c}
         \norm{E_t^\text{VAD} - \mathbf{e}_{w_0}} \\
         \dots \\
         \norm{E_t^\text{VAD} - \mathbf{e}_{w_{|V|}}} 
    \end{array}\right]
\end{equation*}

In the following, we set a constant $\lambda=1$ after generating the first $\text{max}_\text{length}/{2}$ words, as this setting ensures that the first generated words carry the right emotional content, while not sacrificing the grammatical correctness of the whole response. This leads to an improvement in performance.

\subsection{Affective Inference} 
\label{ssec:reranking}
The methods described in the previous sections aim to improve the seq2seq training/sampling procedure. We hypothesize that a good inference strategy is crucial for generating diverse and emotion-specific responses. As \newcite{mmi} suggest, traditional objective functions, \ie likelihood of a response given an input, can be improved by using an $N$-best list and \ac{MMI} during inference. We build upon this idea; our hypothesis is that by generating $B$ diverse sequences and re-ranking the responses, we are more likely to infer one best emotion-specific response. The $B$-best list is found using Beam Search of size $B$ with length normalization.

In the \ac{MMI}-bidi setting, \newcite{mmi} rank all responses found during beam search based on a score calculated as:
\begin{equation} \label{eq:mmi}
\mathcal{R}_\text{final} = \argmax_{\mathcal{R}_{C}} {p(\mathcal{R}_{C}|\mathcal{S}) + \alpha p(\mathcal{S}|\mathcal{R}_{C}) + \beta|\mathcal{R}_{C}|}\text{,}
\end{equation}
where $p(\mathcal{S}|\mathcal{R}_{C})$ is a model with the same architecture as $p(\mathcal{R}_{C}|\mathcal{S})$ trained on reversed prompt-response pairs (T-S), and $|R_{C}|$ is the length of the candidate response, $R_{C}$. We modify this objective in the following form:
\begin{equation} \label{eq:affective_mmi}
\begin{split}
\mathcal{R}_\text{final} = \argmax_{\mathcal{R}_{C}} p(\mathcal{R}_{C}|\mathcal{S},E_0) + \alpha p(\mathcal{S}|\mathcal{R}_{C}) \\+ \beta |\mathcal{R}_{C}| - \gamma \norm{E_{\mathcal{R}_{C}} - E_{0}}\text{,}
\end{split}
\end{equation}
where the last term penalizes the deviation of the emotional content, $E_{\mathcal{R}_C}$, of the candidate response, $\mathcal{R}_C$, from the desired emotional content, $E_0$. The task is to find optimal values of parameters $\alpha$, $\beta$ and $\gamma$, which give the best responses in terms of grammatical correctness, diversity ($\alpha$, $\beta$) and emotional content ($\gamma$) (see \textsection \ref{sec:Results} and \textsection \ref{sec:human-in-the-loop}). 


\section{Model Training} \label{sec:Experiments}
In this section, we describe corpora used for training, the baseline models and the training procedure for the models presented in \sect\ref{sec:General_Model}.

\begin{table*}[ht]
\normalsize\addtolength{\tabcolsep}{-4pt}
  \centering
      \begin{tabular}{r@{}cccccccc} \cmidrule{3-9}
          &&Model & C distinct-1 & C distinct-2 & OS distinct-1 & OS distinct-2 & C BLEU & OS BLEU \\ \cmidrule{3-9}
          \multirow{7}{*}{\rotatebox{90}{No re-rank}}&\multirow{7}{*}{$\left\{\vphantom{\begin{tabular}{c}3\\3\\3\\3\\3\\3\\3\end{tabular}}\right.$} & Baseline & 0.0305 & 0.1402 & 0.0175 & 0.1205 & 0.0096 & 0.094 \\
          && ECM & 0.0310 & 0.1412 & 0.0180 & 0.1263 & 0.0099 & 0.099 \\
          && SEE & 0.0272 & 0.1331  & 0.0170 & 0.1100 & 0.0110 & 0.093 \\ 
          && SED & 0.0303 & 0.1502 & 0.0189 & 0.1231 & 0.0128 & 0.103 \\
          && WI & 0.0316 & 0.1480  & 0.0175 & 0.1235  & 0.0129 & 0.100 \\
          && WE &  0.0310 & 0.1400  & 0.0195 & 0.1302 & 0.0098 & 0.095 \\
          && WI + WE & \bf 0.0342 & \bf 0.1530 & \bf0.0198 & \bf0.1300 & \bf0.0108 & \bf0.105 \\
          &&& \bf (+12.1\%) & \bf(+9.1\%) & \bf (+13.1\%) & \bf (+7.9\%) & \bf (+12.5\%) & \bf (+11.7\%) \\ \cmidrule{3-9}
          \multirow{3}{*}{\rotatebox{90}{Re-rank}}&\multirow{3}{*}{$\left\{\vphantom{\begin{tabular}{c}3\\3\\3\end{tabular}}\right.$}&MMI$_{baseline}$ & 0.0379 & 0.1473& 0.0200 & 0.1403 & 0.0130 & 0.105 \\
          &&EMOTICONS$_{\gamma=0}$ & \bf 0.0406 & \bf 0.2030& \bf0.0305 & \bf0.1431 & \bf0.0140 & \bf0.110 \\
          &&& \bf (+7.1\%) & \bf (+37.8\%) & \bf (+52.5\%) & \bf (+2.0\%) & \bf (+7.7\%) & \bf (+4.8\%) \\ \cmidrule{3-9}
      \end{tabular}
  \caption{\textbf{Quantitative results.} Results for all proposed models trained on \ac{cornell} (C) and OpenSubtitles (OS). distinct-1 and distinct-2 count the number of distinct unigrams and bigrams, respectively, normalized by the total number of generated tokens in 200 candidate responses. The performance boost is computed with respect to the vanilla seq2seq model.}
  \label{tab:results}
  \vspace{-1.2em}
\end{table*}

\label{sec:dataset}
\subsection{Corpora}

\textbf{\ac{cornell}} contains around 10K movie characters and around 220K dialogues \cite{cornell}.
\\\textbf{OpenSubtitles2018} is a collection of translated movie subtitles with 3.35G sentence fragments \cite{opensubtitles}. It has been filtered to get pairs of consecutive sequences (containing between 5 and 30 words), with respective timestamps within an interval of 5 seconds, that are part of a conversation of at least 4 turns. The filtered dataset contains 2.5M utterances.
\\\textbf{Preprocessing} Each dataset is tokenized using the spaCy\footnote{https://spacy.io} tokenizer, converted to lowercase, and non-ASCII symbols are removed. To restrain the vocabulary size and correct the typos, we use a default vocabulary of fixed size 42K words from spaCy. Each word in the dataset is then compared with the vocabulary using the difflib library\footnote{https://docs.python.org/3/library/difflib.html} in Python (algorithm based on the Levenshtein distance), and mapped to the most similar word in the vocabulary. If no word with more than 90\% of similarity is found, the word is considered a rare word or a typo, and is mapped to the out-of-vocabulary (OOV) word. For \ac{cornell}, less than $1\%$ of the unigrams are OOV.

\subsection{Affective Dictionary}\label{sec:affective-dictionary} The \ac{VAD} lexicon may not have all the words in the vocabulary. Based on the word similarity (using difflib library), each word of the vocabulary is assigned a \ac{VAD} value of the most similar word in the VAD lexicon. If no word with more than 90\% of similarity is found, a ``neutral'' \ac{VAD} value ($\textbf{v} = [0.5,0.5,0.5]$) is assigned.

\subsection{Baselines}
We compare our work to two different baselines: a vanilla seq2seq and the ECM introduced by \newcite{aaai_emotion}. For the external memory we use our affective dictionary and train the model using the default parameters provided by authors.

\subsection{Training Details}
All the hyper-parameters have been optimized on the validation set using BLEU score \cite{bleu}. For the encoder, we use two-layer bidirectional \acp{GRU} (hidden size of $256$). The final hidden states from both directions are concatenated and fed as an input to the decoder of one-layer uni-directional \acp{GRU} (hidden size of $512$). The embedding layer is initialized with pre-trained word vectors of size $300$ \cite{subword-embeddings}, trained with subword information (on Wikipedia 2017, UMBC web-base corpus and statmt.org news dataset), and updated during training. We use ADAM optimizer \cite{adam} with a learning rate of $0.001$ for learning $p(\mathcal{R}_C|\mathcal{S}, E_0)$ (resp. $0.01$ for $p(\mathcal{S}|\mathcal{R}_C)$), which is updated by using a scheduler with a patience of $20$ epochs and a decreasing rate of $0.5$. The gradient norm is clipped to $5.0$, weight decay is set to $1e^{-5}$, and dropout \cite{dropout} is set to $0.2$. The maximum sequence length is set to $20$ for \ac{cornell} and to $30$ for OpenSubtitles. The models have been trained on $94\%$, validated on $1\%$, and tested on $5\%$ of the data.

\section{Quantitative Evaluation for Model Selection} \label{sec:Results} 

To evaluate language models, we use BLEU score (computed using $1$- to $4$-grams), as it has been shown to correlate well with human judgment \cite{correlation}. Perplexity does not provide a fair comparison across the models: during the training of the baseline seq2seq model, we minimize the cross entropy loss (logarithm of perplexity), whereas in other models (\eg \ac{WI}) we aim to minimize a different loss not directly related to perplexity (cross entropy extended with the affective regularizer). Having more diverse responses makes the affective re-ranking more efficient, to evaluate diversity we count the number of distinct unigrams (distinct-1) and bigrams (distinct-2), normalized by the total number of generated tokens.

The performance of different models introduced in \sect\ref{sec:General_Model} are presented in Table~\ref{tab:results}. MMI$_\text{bas.}$ refers to a system that re-ranks responses based on Equation~\ref{eq:mmi}, where both $p(\mathcal{R}_{C}|\mathcal{S})$ and $p(\mathcal{S}|\mathcal{R}_{C})$ are baseline seq2seq models. \ac{EMOTICONS} is a system based on Equation~\ref{eq:affective_mmi}, where  $p(\mathcal{R}_{C}|\mathcal{S},E_0)$ is computed using a composition of Word-Level Implicit Model (\ac{WI}) and Word-Level Explicit Model (\ac{WE}), and $p(\mathcal{S}|\mathcal{R}_{C})$ is computed using \ac{WI} (as we are not interested in explicitly using the input emotion). We optimize $\alpha$ and $\beta$ on the validation set using BLEU score, since \newcite{mmi} have shown that adding \ac{MMI} during inference improves the BLEU score. We set $\gamma = 0$ and find optimal values $\alpha_{opt}=50.0$ and $\beta_{opt}=0.001$ using grid search.

Improving BLEU score and diversity was not the goal of our work, but the observed improvement (after adding emotions) shows that the different systems are able to extract and use emotional patterns to improve the general language model.

\subsection{Response Diversity}
From Table~\ref{tab:results}, we observe that for both \ac{cornell} and OpenSubtitles datasets, \ac{SED}, \ac{WI}, and \ac{WE} models outperform the vanilla seq2seq and the ECM  for at least one of the two distinct measures. 
\ac{SEE} has the worst performance overall and does not compete with either the baseline, nor with \ac{SED}. This is expected according to the results reported by \newcite{emotion-token}. It seems that the model is not able to capture the information carried by the additional emotion embedding token -- it is treated as just one additional word among 20 others. \ac{SED} makes better use of the emotion information, as it is used at each time step during decoding. In addition, it is more natural to use these features during the decoding, since the emotion embedding represents the desired emotion of the response. The combination of \ac{WI} and \ac{WE} performs best in terms of distinct-1 and distinct-2 measures among all models without re-ranking, yielding an improvement of up to $13.1\%$. It suggests that the word level emotion models suit the seq2seq architecture better. During training, both models are encouraged not only to match the target words, but also to promote less frequent words that are close to the target words in terms of \ac{VAD} values (affective regularizer and affective sampling), fostering the model to generate more diverse responses.

As expected, by adding MMI, we observe an improvement in diversity, but the relative improvement for OpenSubtitles (MMI$_\text{bas.}$) is smaller than the one reported by \newcite{mmi}. This could originate from the different data filtering and beam search strategy, and the fact the hyper-parameter optimization has been performed on \ac{cornell}. \ac{EMOTICONS} is a combination of $\ac{WI} + \ac{WE}$ (best performing model) for $p(\mathcal{R}_{C}|\mathcal{S}, E_0)$ and $\ac{WI}$ for $p(\mathcal{S}|\mathcal{R}_{C})$, it is better than MMI$_\text{bas.}$ (up to $52.5\%$ gain in distinct-1).

It is worth noting that we observe higher scores in terms of diversity for the reversed model $p(\mathcal{S}|\mathcal{R}_{C})$ compared to the normal model $p(\mathcal{R}_{C}|\mathcal{S}, E_0)$, while training on \ac{cornell}. We can explain this using the data distribution: distinct-2 is higher for the questions than for the answers ($0.167$ and $0.154$ for \ac{cornell}, respectively).

\subsection{Response Quality}
Table~\ref{tab:results} shows that, in general, introducing emotional features into the process of generating responses does not reduce the BLEU score. To reduce the potential negative impact of choosing inappropriate first words in the sequence, we compute the BLEU score on the result of beam search of size 200. For example, if the first word is ``I'', the seq2seq models tend to generate a response ``I don't know'' with high probability, due to the high number of appearances of such terms in the training set. In certain cases, like \ac{WI} and \ac{SED}, we observe an improvement. Such an improvement is expected, since our model takes into account additional (affective) information from the target sequence during response generation.
\begin{figure}
\centering
\includegraphics[width=\linewidth]{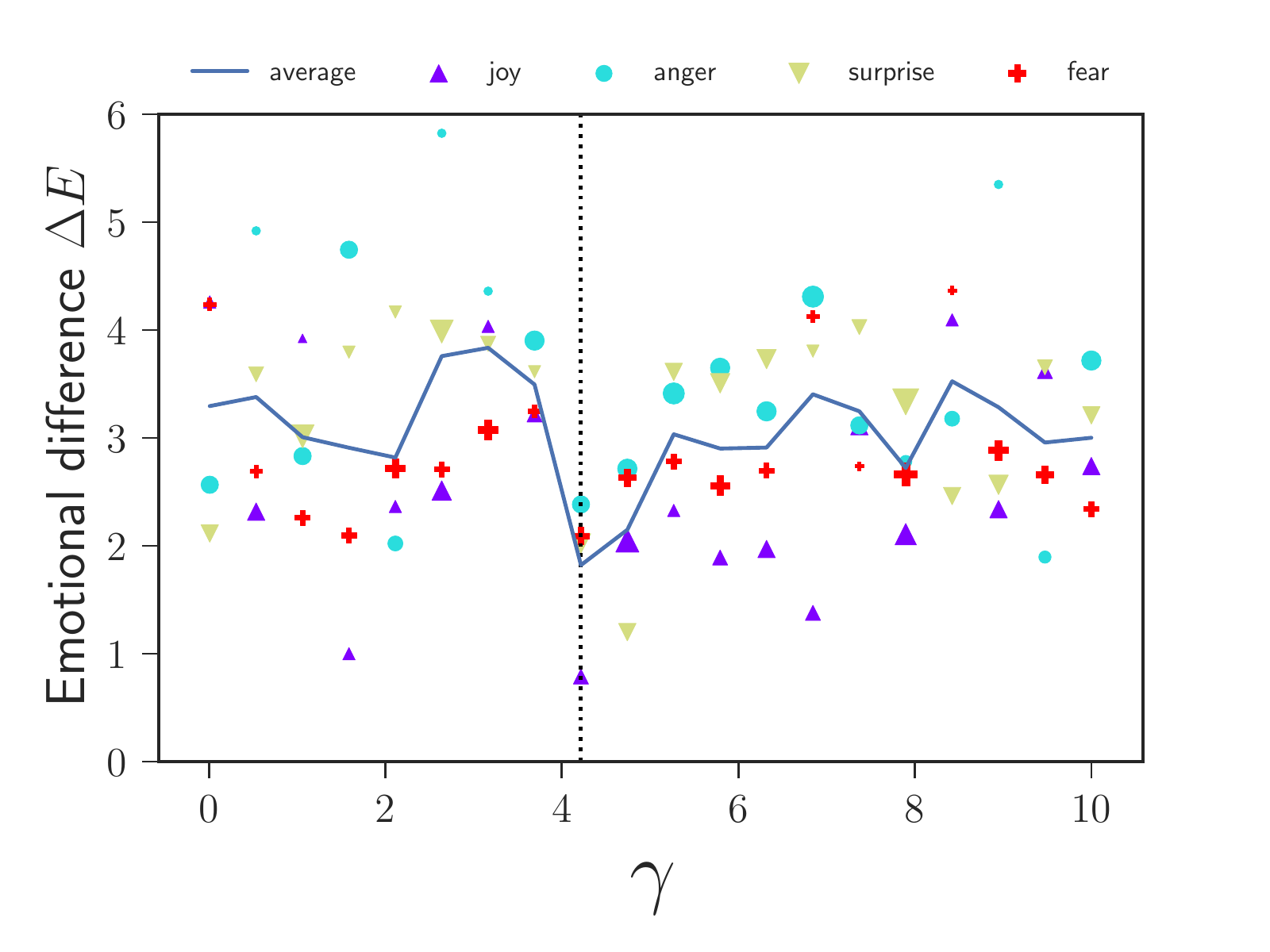}
\caption{\textbf{Hyper-parameter optimization.} For different values of $\gamma$, users assign a face to the generated response. Each point represents an average $\Delta E$ of annotations for each emotion. $\Delta E$ is the difference between the \ac{VAD} representation of the face assigned by the user and the desired emotion for the response. The size is proportional to the number of collected annotations. $\Delta E$ is at a minimum for $\gamma_\text{opt}=4.2$.}
\label{fig:parameter_optimization}
\vspace{-1.3em}
\end{figure}

\section{Human-in-the-Loop Hyper-Parameter Estimation} \label{sec:human-in-the-loop}
The quantitative evaluation shows that \ac{EMOTICONS} outperforms the baseline while adding the emotional features during response generation. The re-ranking phase did not take into account the affective term ($\gamma=0$ in Equation~\ref{eq:affective_mmi}). Setting a different value would not necessarily improve any of the available metrics (\eg BLEU score, diversity), as they do not explicitly take into account affective content in their definition. In this section, we describe an optimization procedure, relying on human judgment, for finding the optimal value of $\gamma$.

\subsection{Experiment Description} We asked annotators to evaluate (using an AffectButton) the generated  responses. We use AffectButton \cite{affectbutton}, a reliable affective tool for assigning emotions, which, to our knowledge, has never been used for estimating the emotional content of the generated responses. In our experiment, the AffectButton lets users choose a facial expression from a continuous space (see Figure \ref{fig:final_figures}), that best matches the emotional state associated with the sequence, which is then mapped into the \ac{VAD} space.
In order to conduct the experiment, we chose a pool of $12$ annotators, who annotated a total of $400$ sequences. The prompts were randomly chosen from the test set of \ac{cornell}, among the $200$ sequences that create the most diverse responses in terms of distinct-2. The more diverse the responses are, the more likely we are to select a response carrying a desired emotion. The responses for the prompts were generated using \ac{EMOTICONS} where the target emotion was either fear, anger, joy, or surprise; the four corners of the AffectButton. $\gamma$ was randomly chosen among $20$ uniformly sampled values in $[0,10]$.

\subsection{Experiment Results}
In \fig~\ref{fig:parameter_optimization}, we present the difference between the VAD value according to the face assigned by the user, and the desired emotion for the response. The average curve presents a global minimum at $\gamma_{opt} = 4.2$. The system does not perform equally well at generating different emotions according to the human judgment. On average, we observe lower values for joy compared to anger in \fig~\ref{fig:parameter_optimization}. This phenomenon is expected, as in the re-ranking process $E_{\mathcal{R}_\text{C}}$ is estimated using the emotion classifier \cite{emotion-classifier} which detects joy more accurately than anger ($77\%$ versus $57\%$), surprise ($62\%$) and fear ($69\%$).


\section{Qualitative Evaluation}
\label{sec:User_Study}
In this section, we qualitatively evaluate the emotional content and correctness of the responses generated by EMOTICONS$_{\gamma=\gamma_{opt}}$ compared to the ones from MMI$_\text{bas.}$ through a user study. It consists of three different experiments which measure grammatical correctness, user preference, and emotional appropriateness. For all experiments, we chose prompts from the test set of \ac{cornell}, for which the most diverse responses were created by MMI$_\text{bas.}$ in terms of distinct-2. We test \ac{EMOTICONS} by generating responses according to four emotions: fear, anger, joy, and surprise (beam size of 200).

\subsection{Grammatical Correctness} \label{sec:grammar}
In this experiment, we used 40 prompts. For each prompt, we generated 5 sentences (4 for \ac{EMOTICONS}, and 1 for MMI$_\text{bas.}$) that were presented in a random order to 3 native English speakers. They assigned either 0 (sentence grammatically incorrect), or 1 (sentence grammatically correct) for all sentences. To measure the agreement across annotators, we calculate Fleiss' $\kappa=0.4128$, which corresponds to ``moderate agreement''. Our model does not substantially sacrifice the grammatical correctness of the responses (see Table~\ref{tab:gramar_exp}).
\begin{table}
\normalsize\addtolength{\tabcolsep}{-4pt}
    \centering
    \begin{tabular}{ccccc}
    \cmidrule{2-5}
        &Model & Grammatical &\multicolumn{2}{c}{User Preference}\\
        &&Correctness&Total & Majority Vote \\\cmidrule{2-5}
        &MMI$_\text{bas.}$ & 83 \% &39&8\\\cmidrule{2-5}
        \multirow{4}{*}{\rotatebox{90}{\scriptsize EMOTICONS}}&Fear & 82 \% &\multirow{4}{*}{96}&\multirow{4}{*}{37}\\
        &Anger & 80 \% &&\\
        &Joy & 84 \% &&\\
        &Surprise & 79 \% &&\\\cmidrule{2-5}
    \end{tabular}
    \caption{\textbf{User study results.}
    \textit{Grammatical Correctness} shows the ratio of grammatically correct sentences among all generated responses, whereas \textit{User Preference} shows the number of times each model was preferred by the users.}
    \label{tab:gramar_exp}
    \vspace{-1.0em}
\end{table}
\begin{figure}
\centering
\begin{subfigure}{.09\textwidth}
  \centering
  \includegraphics[width=\linewidth]{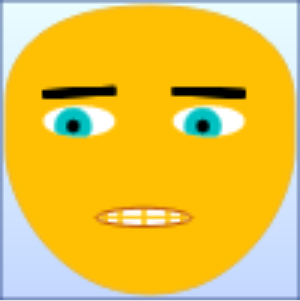}
    \caption{Bas.}
  \label{fig:baseline}
\end{subfigure}
\begin{subfigure}{.09\textwidth}
  \centering
  \includegraphics[width=\linewidth]{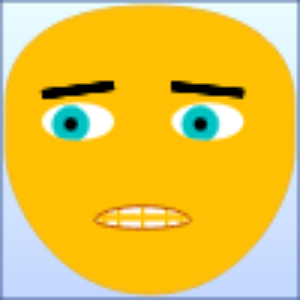}
    \caption{Fear}
  \label{fig:fear}
\end{subfigure}
\begin{subfigure}{.09\textwidth}
  \centering
  \includegraphics[width=\linewidth]{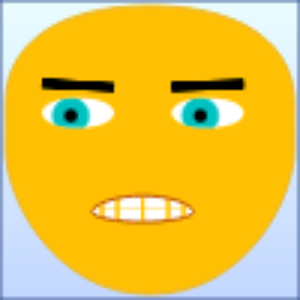}
    \caption{Anger}
  \label{fig:anger}
\end{subfigure}
\begin{subfigure}{.09\textwidth}
  \centering
  \includegraphics[width=\linewidth]{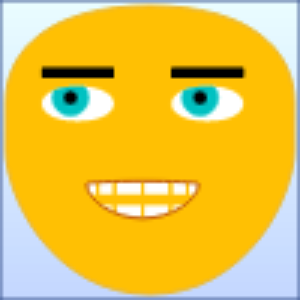}
    \caption{Joy}
  \label{fig:joy}
\end{subfigure}
\begin{subfigure}{.09\textwidth}
  \centering
  \includegraphics[width=\linewidth]{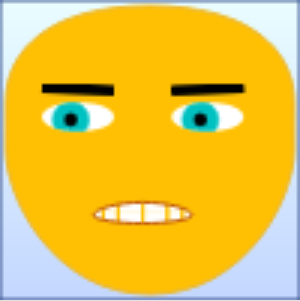}
    \caption{Surprise}
  \label{fig:surprise}
\end{subfigure}
\caption{\textbf{Emotional faces.} AffectButton presents faces according to average VAD vectors (in parentheses) obtained for the (a)~MMI$_\text{bas.}$ $\left([0.47, 0.98, 0.36]\right)$, and for the four EMOTICONS models with different target emotions: (b)~Fear $\left([0.2, 0.95, 0.38]\right)$, (c)~Anger $\left([0.54, 0.92, 0.65]\right)$, (d)~Joy $\left([0.68, 0.97, 0.66]\right)$, and (e)~Surprise $\left([0.37, 0.97, 0.52]\right)$.}
\label{fig:final_figures}
\vspace{-1.0em}
\end{figure}
\subsection{User Preference} \label{sec:user-preference}
In this setting, we quantify how likely the user is going to prefer the response generated by \ac{EMOTICONS} compared to the one generated by MMI$_\text{bas.}$. We asked $18$ annotators to choose their favorite response to the input query among eight proposed answers (top four responses coming from the MMI$_{bas}$ and 4 coming from \ac{EMOTICONS} with the four different target emotions). Each of 45 sentences were annotated by three different annotators. Results of the experiment (Table~\ref{tab:gramar_exp}) indicate that users strongly prefer \ac{EMOTICONS} over MMI$_\text{bas.}$.
\subsection{Emotional Appropriateness} \label{sec:emotion}In this experiment, we show that our model is able to generate emotions in a controlled manner. For each of the 5 models, $22$ users assign a face via the AffectButton. We generate responses for 120 different prompts. We keep the responses that were annotated with a \ac{VAD} vector with the norm greater than $2$, corresponding to those expressing strong emotions. We compute the average \ac{VAD} vectors for the annotated sequences for each model, with corresponding AffectButton faces (\fig~\ref{fig:final_figures}). The majority of user-assigned faces have a high arousal value, which can be explained by the fact that users tend to click in one of the four corners of the AffectButton. The majority of the faces represent an accurate portrayal of the desired emotion. The poor performance of \ac{EMOTICONS} at expressing surprise comes from the fact that (1) users often mismatch surprise with joy, leading to a neutral dominance value, and (2) surprise is one of the most difficult emotions to judge (see \sect\ref{sec:human-in-the-loop}).

\section{Conclusion} \label{sec:Conclusions}
We have presented \ac{EMOTICONS}, a system that can generate responses with controlled emotions. The flexibility of the presented solution allows it to be used in any kind of neural architecture as long it fits the encoder-decoder framework.
Currently, \ac{EMOTICONS} does not generate different emotions equally well. Future work could include incorporating contextual information that would help \ac{EMOTICONS} to better capture emotional content.

\section*{Acknowledgments}
We would like to thank anonymous reviewers for their insightful comments. 
Mubbasir Kapadia has been funded in part by NSF IIS-1703883, NSF S\&AS-1723869, and DARPA SocialSim-W911NF-17-C-0098.

\bibliography{naaclhlt2019}
\bibliographystyle{acl_natbib}

\end{document}